\documentclass[a4paper,conference]{IEEEtran}
\usepackage{lipsum}
\usepackage{graphicx}
\usepackage{comment}
\usepackage{amsmath,amssymb} 
\usepackage{color}
\usepackage{amsmath,amssymb,amsfonts}
\usepackage[ruled,vlined, linesnumbered]{algorithm2e}
\usepackage{cite}
\usepackage{textcomp}
\usepackage{xcolor}
\usepackage{multirow}
\usepackage{wrapfig,lipsum,booktabs}
\usepackage{makecell}
\usepackage{tabularx}

\ifCLASSINFOpdf
\else
\fi
\begin{document}
%
\title{Activation Density driven Efficient Pruning in Training}





\author{\IEEEauthorblockN{Timothy Foldy-Porto}
\IEEEauthorblockA{Department of Electrical Engineering\\ Yale University
}
\and
\IEEEauthorblockN{Yeshwanth Venkatesha}
\IEEEauthorblockA{Department of Electrical Engineering\\ Yale University
}
\and
\IEEEauthorblockN{Priyadarshini Panda
}
\IEEEauthorblockA{Department of Electrical Engineering\\ Yale University
}
}

\maketitle




\begin{abstract}
Neural network pruning with suitable retraining can yield networks with considerably fewer parameters than the original with comparable degrees of accuracy. Typical pruning methods require large, fully trained networks as a starting point from which they perform a time-intensive iterative pruning and retraining procedure to regain the original accuracy. We propose a novel pruning method that prunes a network real-time during training, reducing the overall training time to achieve an efficient compressed network. We introduce an activation density based analysis to identify the optimal relative sizing or compression for each layer of the network. Our method is architecture agnostic, allowing it to be employed on a wide variety of systems. For VGG-19 and ResNet18 on CIFAR-10, CIFAR-100, and TinyImageNet, we obtain exceedingly sparse networks (up to $200 \times$ reduction in parameters and over $60 \times$ reduction in inference compute operations in the best case) with accuracy comparable to the baseline network. By reducing the network size periodically during training, we achieve total training times that are shorter than those of previously proposed pruning methods. Furthermore, training compressed networks at different epochs with our proposed method yields considerable reduction in training compute complexity ($1.6\times$ to $3.2\times$ lower) at near iso-accuracy as compared to a baseline network trained entirely from scratch. 
\end{abstract}

%
\IEEEpeerreviewmaketitle

\section{Introduction}

Deep learning has proliferated in the past decade making its way into numerous applications. Not only has it captured the public's imagination as a candidate for the development of intelligent systems, but it has achieved high accuracy on difficult datasets. Particularly, deep networks have performed well on computer vision tasks and natural language processing \cite{simonyan2014very}, \cite{he2016deep}. Part of their success has been attributed to the networks' depths---typical deep neural networks can comprise of hundreds of layers---but this comes with the cost of having a huge number of trainable parameters \cite{liu2018rethinking}.

The concept of \textit{network pruning}---systematically reducing the number of parameters in a given network configuration---has been around since the early 1990s \cite{lecun1990optimal}, but only recently it has begun to receive widespread attention. Over the past five years, many network pruning strategies have been proposed and the motivations for pruning have been explored \cite{liu2018rethinking, garg2018low, han2015learning, han2015deep, frankle2018lottery, zhou2016less, denton2014exploiting, liu2018dynamic}. Iandola et al. \cite{iandola2016squeezenet} have identified three ways in which pruned architectures are superior to the networks from which they were created: they are more efficiently trained on distributed systems, their smaller model size makes them easier to send to new clients (a self-driving car, for example), and they are more suited for deployment on edge devices such as mobile phones or embedded processors. 

Most network pruning algorithms that have been proposed follow the same structure:
\begin{itemize}
    \item Train a large network to a high degree of accuracy.
    \item Prune the model architecture while preserving the required set of weights.
    \item Fine-tune the pruned model to regain accuracy.
\end{itemize} 
The need for a large fully trained network slows down the entire pruning process, since it is often computationally expensive and time consuming to train large models. The necessity of having a fully trained network as starting point comes from the assumption that the pruned network will train better if it is initialized using the weights of its high-performing dense network than if it is randomly initialized. Furthermore, most pruning strategies rely on the significance of the weight values (say, L1 norm \cite{liu2017learning}) to decide if the connection should be pruned or not. Thus, a pre-trained network is necessary in such cases to assign significance. 

However, Liu et al. \cite{liu2018rethinking} have shown that it is not necessary to preserve weight values when moving from a large model to a smaller model. They found a negligible discrepancy between fine-tuning a pruned model and training that same model from randomly initialized weights. This implies that, in order to successfully reduce the size of a large network without a significant loss in accuracy, it is sufficient to find only the optimal architecture of the pruned network. Finding the significant weights to prune or initialize a compressed network (as in \cite{franklestabilizing}, \cite{han2015learning}, \cite{roy2020pruning} and \cite{iandola2016squeezenet}) is not always necessary. In light of this, we make the following key contributions:
\begin{itemize}
    \item We propose an in-training pruning method that analyzes the network performance in real-time and optimizes the architecture throughout the training process. Unlike many prior pruning methods, the method presented here does not require a fully pre-trained network. Rather, our approach progressively prunes a model in a structured manner during the training process.
    \item We introduce a simple, reliable, and computationally-efficient metric of \textit{activation energy (AE)}, upon which our pruning algorithm relies. AE is a significance-agnostic metric that determines the optimal layer-specific sizing of a network without any dependence on which weights to keep or prune. 
    \item We demonstrate considerable reduction in compute operations (OPS) and training complexity exceeding that of other pruning algorithms for prevailing benchmark datasets: CIFAR-10, CIFAR-100 and TinyImageNet.
\end{itemize}

\section{Related Work}

Much work has been done in the past few years with regards to network compression. Some of the earlier approaches focus on the idea of compressing a pre-trained network according to some salience criteria. Denton et al. \cite{denton2014exploiting} apply singular value decomposition to a pre-trained convolutional network. Han et al. \cite{han2015learning} identify weights that are below a certain threshold and replace them with zeros to produce a sparse network, which is then fine-tuned for a few iterations to produce the final pruned network. Han et al. \cite{han2015deep} also introduced Deep Compression, a technique that combines pruning methods with quantization and Huffman coding to achieve substantial improvements in compute and energy efficiency. Other methods prune on the scale of channels or layers. Wen et al. \cite{wen2016learning} developed Structured Sparsity Learning (SSL), which regularizes the architecture of a pre-trained model to achieve speedup in inference. Zhou et al. \cite{zhou2016less} enforce channel-level sparsity during the training process.

More recently, the need for pre-trained networks has been questioned. Liu et al. \cite{liu2018rethinking} advocate a rethinking of structured pruning techniques, demonstrating that transferring weights from pre-trained networks to pruned networks is not as beneficial as is traditionally thought. Concurrently, Frankle \& Carbin \cite{frankle2018lottery} introduced the ``lottery ticket hypothesis", which theorizes the existence of small sub-networks (`winning tickets') that train faster and to the same degree of accuracy as the larger networks in which they were found. They propose iterative magnitude pruning that finds winning tickets that are $90-95\%$ less dense (in terms of parameters) while maintaining a competitive accuracy compared to the original dense network. The authors in \cite{lee2018snip} use a connection sensitivity based approach to prune the network in a single step at the time of initialization. Roy et al. \cite{roy2020pruning} use a dynamic pruning while training approach, but use significant weight assignment metric such as, L1 normalization to perform the pruning.


Both parameter reduction and OPS reduction have significant benefits in deploying a neural network. In practice, however, the two terms have subtly different implications. The parameters of a network are what occupy memory space; the number of OPS, while closely related to the number of parameters, is an indicator of the compute effort of passing an input through the network. OPS reduction with pruning consequently improves the energy efficiency in hardware accelerators that take advantage of sparsity \cite{parashar2017scnn, page2017sparcnet, zhang2016cambricon, han2017ese}. As the translation from number of OPS to energy efficiency is not straight forward and heavily dependent on hardware platform \cite{8335698, DBLP:journals/corr/YangCS16a}, detailed energy profiling of different hardware platforms is outside the scope of this work. We use the number of OPS as a proxy for compute efficiency. \textbf{In this paper, we focus on reducing the compute cost of networks during both training and inference. Compared to other pruning methods, we demonstrate greater OPS reduction and a lower compute cost for training at iso-accuracy with respect to baseline model.} 


\section{Pruning in Training Methodology}\label{pruningtrainingmethod}

\begin{figure*}
    \centering
    \includegraphics[width=0.85\textwidth]{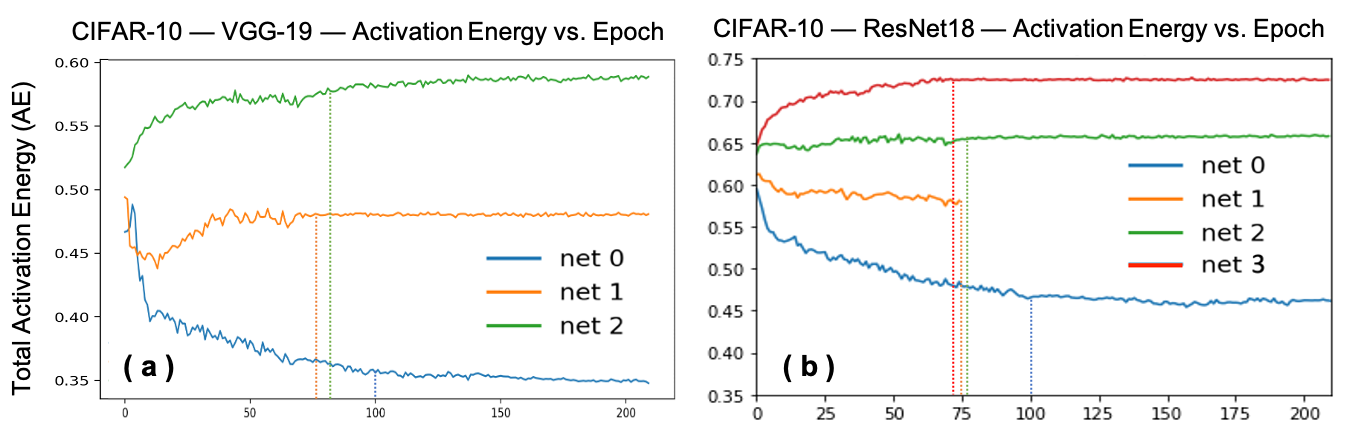}
    \caption{Activation energy (AE) per epoch for successively pruned networks. $net 0$ refers to the baseline unpruned network $net 1$ refers to the pruned version of $net 0$, and so on. Vertical lines represent the points in the training processes which met pruning criteria ($\rho$), as described in Section 3.1. All networks are trained on CIFAR-10. Total AE is calculated by summing the layer-wise AEs across all layers. In (a), $net 0$ is VGG-19. In (b), $net 0$ is ResNet18.}
    \label{fig_1}
    \vspace{-3mm}
\end{figure*}

Our method is motivated by a key observation that, for a randomly initialized network of sufficient initial size, the total density of non-zero activations in the network (note, total density calculated across all layers) decreases during the training process with increasing epochs (see $net 0$ graph in Fig. \ref{fig_1}). The density (also defined as Activation Energy (AE)) is calculated as
\begin{equation}
    AE = \frac{\# nonzero\hspace{2mm} activations}{\#total\hspace{2mm}activations}
\end{equation}
We interpret this decreasing density (AE profile) as evidence of redundant features in the initial network. While there is no \textit{a priori} theoretical basis for this hypothesis, we are able to verify it experimentally by taking the initial network, reducing the number of filters in each layer, retraining the pruned network and observing negligible loss in accuracy. Additionally, the AE profile of the pruned network decreased less over the course of the training process. This finding is exemplified in Fig. \ref{fig_1}. For both VGG-19 and ResNet18, we took the initial network, $net 0$, resized each layer using the proposed AE based pruning method and retrained it to produce the AE profile of the subsequent network, $net 1$. Note, the overall AE of $net1$ is higher than $net0$ in Fig. \ref{fig_1}. This implies that pruning the network based on the AE metric causes more neuronal units to be active which is an indication of less redundancy.  

\begin{figure*}[h!]
    \centering
    \includegraphics[width=0.85\textwidth]{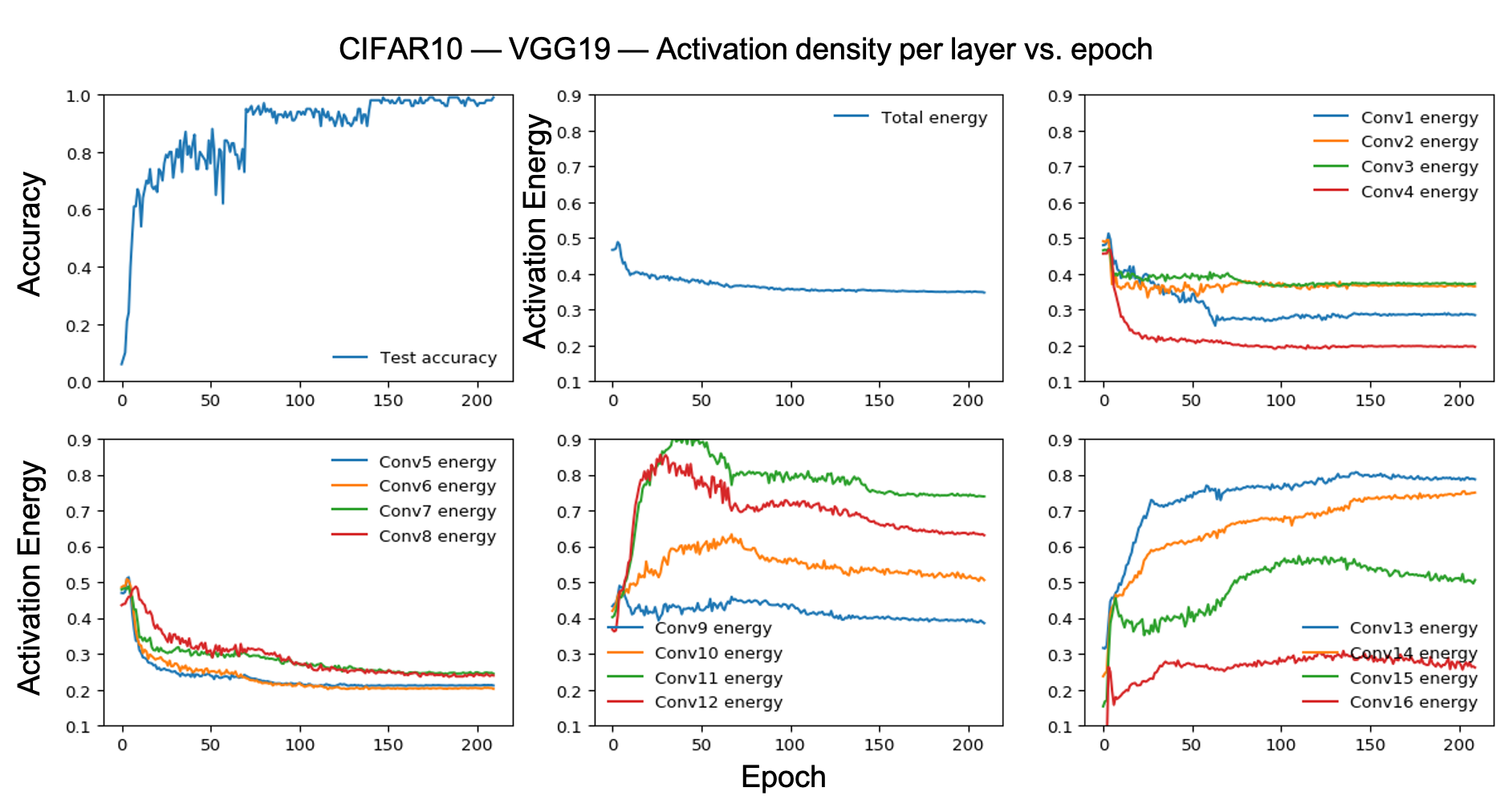}
    \caption{Accuracy, Total AE across all layers and AE per layer with increasing epochs as training proceeds for VGG-19 on CIFAR-10 is shown. We did not include pooling layers or the final fully-connected layer, as there was no additional information present in those layers. The trend of decreasing AE per layer can be seen here. Additionally, we observe a convex activation energy profile for the first eight layers and a concave profile for the second eight layers. }
    \label{fig_2}
    \vspace{-4mm}
\end{figure*}

We also find that the AE trend of every layer in the network varies from one another (see Fig. \ref{fig_2}). Following the above density-redundancy interpretation, we can infer that a convolution layer $l1$ activating only 44\% of its neurons (i.e. $AE_{l1} =0.44$) during a given training period is wasting 56\% of its allotted capacity. In this case, our pruning method  decides that the layer $l1$ only needs to be 44\% of its initial size for the next training round. To summarize, we propose an activation density driven pruning approach that provides a structured way of obtaining the optimal size of each layer of a network real-time during training.


Periodically throughout the training process, we count the number of activations that are non-zero (equivalent to counting positive activations, since the negatives are zeroed out by the Rectified Linear Unit (ReLU) non-linearity \cite{nair2010rectified}) and divide by the number of total activations, yielding an \textit{activation density or energy} (see Eqn. 1). AE of every layer serves as a good metric to decide its compressed size. Essentially, we monitor the AEs of the layers during the training process and prune the layers based on the density at regular training intervals. Note, activation energy and activation density are used interchangeably in the paper.  
For the networks that we tested on, it was sufficient to multiply the activation densities of a given layer $L$ by the layer size at a current training epoch to obtain the layer sizes of the pruned network for the next training epoch as follows:
\begin{equation}
\vspace{-2mm}
    net_{pruned}.layersize[L] = AE[L] \times net_{initital}.layersize[L]
\end{equation}
Here, $layersize$ denotes the number of output channels in a given layer $L$. For instance, if layer $L$ has 64 channels and $AE[L] = 0.5$ for the initial network, layer $L$ has 32 channels in the pruned network.
\vspace{-1mm}
\begin{algorithm}[b!]\label{algorithm_1}

\SetAlgoLined
\textbf{Input:} Training dataset and randomly initialized network $net_{initial}$\\
\textbf{Output:} Trained and pruned network $net_{final}$\\
 $net[0] = net_{initial}$ \\
 \textit{//Note, net[0] can be a large network like \{VGG-19, ResNet18\}}\;
 $epoch = 0$\;
 $index = 0$\;
\While{\textbf{not} stopping ($\delta$) criteria}{ 
$net$ = Randomly Initialized ($net[index]$)\;
\While{\textbf{not}  pruning ($\rho$) criteria}{ train($net,\hspace{2mm}epoch$)\;
\For{L in net.Layers}{
$\#nonzero[L] =\text{count\_nonzero\_activations(L)}$\;
AE[L] = $\frac{\# nonzero[L]}{\#total[L]}$ \;
}
$epoch++$\;
\textit{//Note, we train the network $net[index]$ while monitoring the layer-wise AE till $\rho$ is satisfied.}
}
$index++$\;
\For{L in net.Layers}{
$net[index].LayerSize[L] =$ AE[L] $\times net[index-1].LayerSize[L]$\;
}
\textit{//Note, we prune the network $net[index-1]$ to get the compressed network $net[index]$ based on AE per layer. The pruning continues till $\delta$ is satisfied.}
}
$net_{final} = net[index]$\;
\caption{Activation Density driven Pruning in Training}
\end{algorithm}

Algorithm \ref{algorithm_1} outlines our proposed approach. The key steps of our pruning in training method are: (1) Define an initial network $net[index]$ or $net[0]$ and train it until a pruning criteria $\rho$ is reached (Lines 9-17 in Algorithm 1); (2) Perform AE analysis on the network and obtain the density per layer (Lines 11-14 in Algorithm \ref{algorithm_1}); (3) For each layer, determine the new layer size by multiplying the $net[0]$ layer size by the AE of that layer (Eqn. 2 or Lines 19-21 in Algorithm \ref{algorithm_1}); (4) Define a network $net[1]$ using the newly generated layer sizes. $net[1]$ will be functionally identical to $net[0]$, just smaller; (5) Repeat steps 1-4 on successively pruned networks $net[index++]=net[1], net[2]...$ until a stopping criteria $\delta$ is reached (Lines 7-24 in Algorithm 1). Note, Algorithm \ref{algorithm_1} determines the optimal size of the convolutional layers. The pooling layers of a network are automatically sized based on the preceding convolutional layers. 

We found that it did not make a difference to training convergence or final accuracy whether the pruned network (say, $net[1]$) was initialized with random weights or with learnt weights from the larger network (say, $net[0]$). 
For simplicity, we chose to randomly initialize the network in each pruning round. Note, the activation density or AE only indicates the total number of filters or weights to remove at each pruning step. In this analysis, there is no notion of significant weights that tell us which specific filters to keep or prune. Thus, we randomly remove the filters at every layer based on the density. The fact that our approach is independent of pre-initialization and significant weights implies that network architecture is key for compression, supporting the results of Liu et al. \cite{liu2018rethinking}.  

\subsection{Pruning ($\rho$) Criteria and Stopping ($\delta$) Criteria}

Algorithm 1 describes two different criteria: $\rho$ indicates when to stop the training of the initial network as well as each successively pruned network ($net[index]$); and $\delta$ indicates when to stop the pruning process altogether. Both of these criteria are most easily understood from a visual inspection of the total AE  for each network as the training progresses. Fig. \ref{fig_1} shows a typical graph of AE vs. epoch. In Fig. \ref{fig_1} (a), $net 0$ is a VGG-19 model trained on CIFAR-10. $net 1$ and $net 2$ are successively pruned models. 

AE for $net 0$ decays throughout the entire training process in Fig. \ref{fig_1} (a). However, after a certain point (approximately 100 epochs) it flattens out. Interestingly, we observed a correspondence between this saturation of AE and a saturation of the network accuracy. Both the accuracy and the AE for a given training period can be characterized by two regimes: (1) Before the saturation point, both quantities exhibit noticeable long-term trends as well as short-term fluctuations; (2) After the saturation point, both quantities stabilize and their derivatives seem to approach zero.  Based on this observation, we choose the pruning criteria $\rho$ to be equivalent to this saturation point (100 epochs in $net 0$, for example). Training any further beyond $\rho$ results in minimal change in AE/accuracy and unnecessarily increases the training time. For $net 1$ and $net 2$ in Fig. \ref{fig_1} (a), the saturation point $\rho$ occurs at approximately 70 epochs. Note, in Fig. \ref{fig_1}, we trained each network well past their saturation points for the purpose of demonstrating network behavior in the post-saturation regime. In practice, we stop training each network at their respective $\rho$ values.

The second stopping criteria ($\delta$) is determined on the basis of the overall shape of AE vs. epoch curve for each network. $net 0$ can be characterized as convex, $net 1$ as flat, and $net 2$ as concave for the VGG-19 CIFAR-10 graphs in Fig. \ref{fig_1} (a). We interpreted the convexity of $net 0$ as an indicator of overparameterization or redundancy and that, since AE went down as accuracy went up, the network learned to ignore redundant connections. This means that we can remove those redundancies without significantly damaging the network's performance. $net 2$, in contrast, trended upwards; by our interpretation of the AE metric, this indicates that the network learned to utilize more connections in order to improve its accuracy. Removing connections further will drastically reduce the accuracy of the network. Thus, we decide to stop the overall pruning process when we see an upward-trending or concave AE profile. 

In practice, the $\rho$ criteria is determined by monitoring the total AE during training. If AE does not change a lot ($\Delta AE < 0.001$) between two or more consecutive epochs, we label that as the saturation point and prune the layers based on the layer-wise densities obtained at the end of that particular epoch. The $\delta$ criteria is determined in practice by monitoring the slope of AE vs. training epoch curve and the slope (negative, positive, nearly 0) can tell us if $\delta$ is met.


\begin{figure*}[h!]
    \centering
    \includegraphics[width=0.85\textwidth]{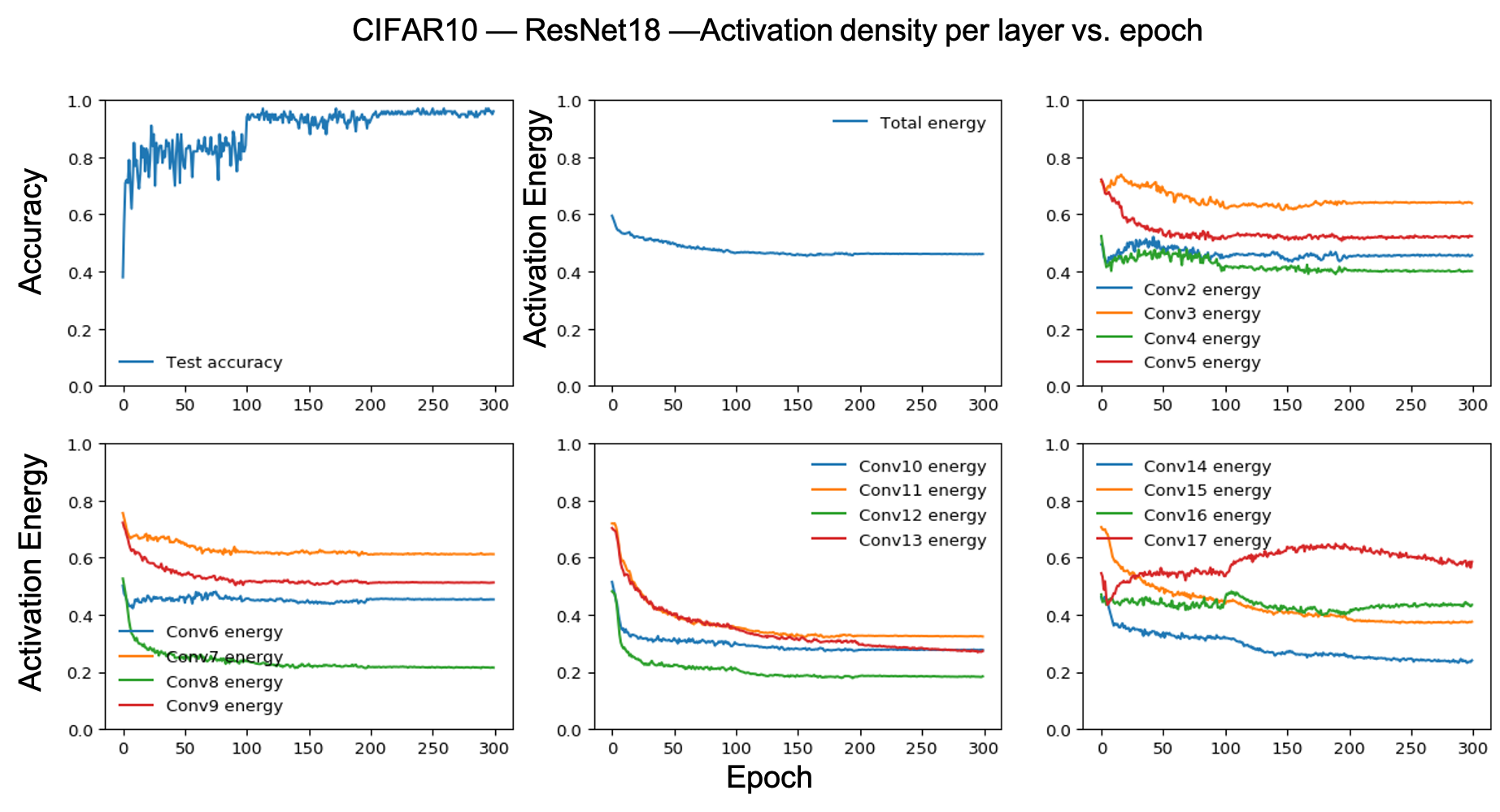}
    \caption{Accuracy, Total AE and AE per layer as training proceeds for ResNet18 on CIFAR-10 is shown. As in the VGG-19 case, the trend of  decreasing AE across every layer can be seen here. Unlike for VGG-19, the concave activation energy profile only emerges in the final layer. }
    \label{fig_5}
    \vspace{-4mm}
\end{figure*}

\begin{table*}[h]
\caption{Summary of results.The final pruned model obtained from our method for each scenario has been \textbf{highlighted}. Training epochs until saturation point $\rho$ is shown. The reported accuracies are from a full training cycle (210 epochs), except for TinyImageNet, which was trained for 60 epochs. Values $>1\times$ denote improvement in OPS, parameters reduction.}
\begin{center}
\resizebox{0.85\linewidth}{!}{
\begin{tabular}{|c|c|c|c|c|c|}
\hline
Network & Configuration & Accuracy & Parameters & OPS & Training  \\
&  &  & reduction & reduction & Epochs $\rho$ \\
\hline
\multicolumn{6}{c}{\textbf{CIFAR-10, ResNet18}}\\
\hline
\textit{net 0} & [ 64, 64, 64, 64, 64, 128, 128, 128, 128, 256, 256, 256, 256, 512, 512, 512, 512 ] & 97 \% & 1$\times$ & 1$\times$ & 100 epochs \\
\hline
\textit{net 1} & [ 34,
29,
41,
25,
33,
58,
78,
27,
65,
71,
83,
46,
69,
120,
191,
219,
288
] & 97 \% & 7.3$\times$ & 6.0$\times$ & 70 epochs \\
\hline
\textbf{\textit{net 2}} & \textbf{[ 21,
16,
30,
10,
22,
24,
47,
9,
39,
26,
48,
12,
39,
41,
85,
63,
188
]} & \textbf{95} \% & \textbf{41.2$\times$} & \textbf{23.2$\times$} & \textbf{70 epochs} \\
\hline
\textit{net 3} & [ 14,
9,
21,
5,
15,
13,
32,
5,
26,
13,
34,
5,
25,
21,
45,
12,
142
] & 91 \% & 199.3$\times$ & 67.1$\times$ & N/A \\
\hline
\multicolumn{6}{c}{\textbf{CIFAR-10, VGG-19}}\\
\hline
\textit{net 0} & [ 64,
64,
128,
128,
256,
256,
256,
256,
512,
512,
512,
512,
512,
512,
512,
512
] & 97 \% & 1$\times$ & 1$\times$ & 100 epochs \\
\hline
\textbf{\textit{net 1}} & \textbf{[ 18,
23,
47,
25,
54,
51,
62,
61,
197,
258,
378,
322,
402,
383,
259,
134
]} & \textbf{94 \%} & \textbf{3.1$\times$} & \textbf{5.6$\times$} & \textbf{70 epochs} \\
\hline
\textit{net 2} & [ 10,
9,
30,
11,
21,
31,
22,
21,
62,
70,
113,
141,
256,
299,
194,
71
] & 93 \% & 10.3$\times$ & 27.4$\times$ & N/A \\
\hline
\multicolumn{6}{c}{\textbf{CIFAR-100, ResNet18}}\\
\hline
\textit{net 0} & [ 64, 64, 64, 64, 64, 128, 128, 128, 128, 256, 256, 256, 256, 512, 512, 512, 512 ] & 81.0 \% & 1$\times$ & 1$\times$ & 25 epochs \\
\hline
\textbf{\textit{net 1}} & \textbf{[ 39,
31,
49,
24,
44,
54,
90,
36,
84,
88,
155,
65,
136,
130,
231,
105,
300
]} & \textbf{79.0 \%} & \textbf{7.6$\times$} & \textbf{5.1$\times$} & \textbf{N/A} \\
\hline
\multicolumn{6}{c}{\textbf{CIFAR-100, VGG-19}}\\
\hline
\textit{net 0} & [ 64,
64,
128,
128,
256,
256,
256,
256,
512,
512,
512,
512,
512,
512,
512,
512
] & 76.0 \% & 1$\times$ & 1$\times$ & 25 epochs \\
\hline
\textbf{\textit{net 1}} & \textbf{[ 34,
23,
51,
30,
63,
63,
73,
82,
210,
285,
333,
357,
317,
259,
181,
106
]} & \textbf{73.0 \%} & \textbf{3.9$\times$} & \textbf{5.3$\times$} & \textbf{N/A} \\
\hline
\multicolumn{6}{c}{\textbf{TinyImageNet, ResNet18}}\\
\hline
\textit{net 0} & [ 64, 64, 64, 64, 64, 128, 128, 128, 128, 256, 256, 256, 256, 512, 512, 512, 512 ] & 51.54 \% & 1$\times$ & 1$\times$ & 25 epochs \\
\hline
\textbf{\textit{net 1}} & \textbf{[ 31,
21,
47,
27,
48,
62,
99,
58,
94,
85,
161,
69,
133,
93,
152,
56,
247
]} & \textbf{50.51 \%} & \textbf{10.6$\times$} & \textbf{4.7$\times$} & \textbf{N/A} \\
\hline
\end{tabular}
}
\label{tab1}
\end{center}
\end{table*}


\subsection{Layer-wise sensitivity to pruning}
While the total AE (in Fig. \ref{fig_1}) of the network provides a convenient and interpretable stopping criteria (that is understood intuitively), the pruning process itself relies only on the layer-wise AE profile. For VGG-19 (see Fig. \ref{fig_2}), we noticed that the layer-wise AE profiles varied greatly: \textit{layers 1-8} exhibited the same concavity seen in the total AE profile, though the scale of the AE profiles tended to decrease towards deeper layers. That is, the AE value of deeper layers, say \textit{layers 7 \& 8}, at the saturation point $\rho$ tended to be lower than the AE value of shallower layers, say \textit{layers 1 \& 2}. \textit{Layers 9-16} exhibited varying degrees of convexity, in opposition to the total AE trend of the network. 


For the first half of the VGG-19 network, AE decreases with layer depth. 
The trend reverses after the network's midpoint (after \textit{layer 8}): AE starts to increase as the network gets deeper. This observation aligned with the discussion of layer-wise pruning in \cite{garg2018low}. The authors in \cite{garg2018low} found using principal component analysis (PCA) that the number of significant dimensions contributing to the variance of the activations decreased past the mid-point of a VGG network. They removed the deeper layers based on this observation, and found a negligible degradation of accuracy. Along similar lines, we also interpreted the reversed AE trend after $layer8$ of VGG-19 to mean that the last layers were no longer identifying increasingly abstract features from the input data and that removing them would have little effect on the network's performance. However, we found that removing whole layers severely degraded the accuracy of the pruned networks. For VGG-19 on CIFAR-10, the accuracy of our first pruned network (with just the first 8 layers intact and pruned based on AE) never exceeded 70\%, even after a full training cycle (210 epochs with learning rate decay). This implies that depth is significant to achieving good training convergence. Furthermore, we found that the overall reduction in inference compute OPS we achieved with AE-based pruning on a VGG-19 network is higher than that of \cite{garg2018low} at iso-accuracy (see Section 4.4), even without removing the latter layers. 

Analysing the layer-wise AE profile (Fig. \ref{fig_5}) of a ResNet18 model trained on CIFAR-10 shows a uniform concave trend across all layers similar to the total AE trend (as in $net 0$ of Fig. \ref{fig_1} (b)) with no reversal at the network's midpoint. To perform network agnostic pruning without limiting the training convergence, we, therefore, use AE as an indicator of compression width per layer and not of the overall depth. 

\section{Results}

We evaluate our AE based pruning in training strategy on two commonly used networks: VGG-19 and ResNet18 for CIFAR-10, CIFAR-100 \cite{krizhevsky2009learning}, and TinyImageNet datasets \cite{tiny}. We imported github models from \cite{github1} for implementing our experiments in PyTorch.  We used similar hyperparameters as \cite{frankle2018lottery} and \cite{github2} to train our models on CIFAR-10/100 and TinyImageNet, respectively.

\subsection{Efficiency analysis}


\begin{figure}[!]
    \centering
    \includegraphics[width=0.85\columnwidth]{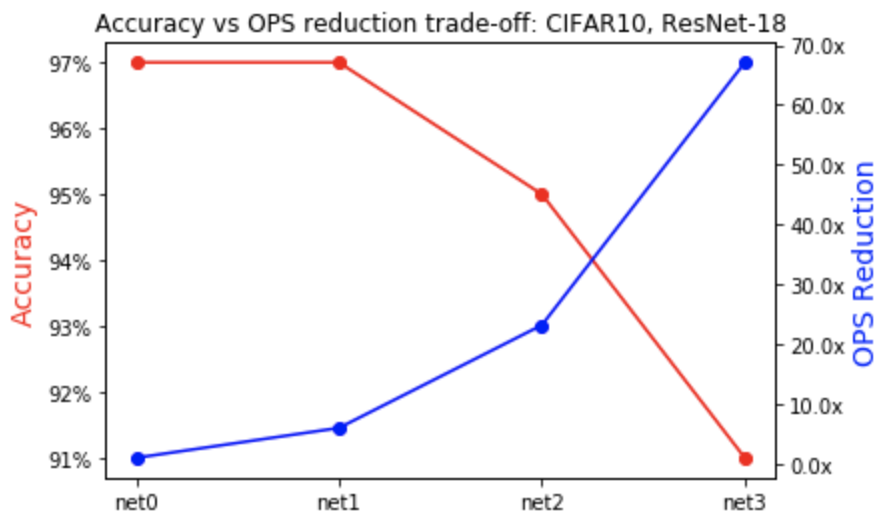}
    \caption{Trade-off between accuracy and OPS reduction with increasing level of pruning.}
    \label{accVsOPS}
    \vspace{-6mm}
\end{figure}

To perform efficiency analysis, we consider \textit{Parameter reduction} and \textit{OPS reduction} as our key metrics. For a particular convolutional layer of a network with $N$ input channels, $M$ output channels, input map size $I\times I$ , weight kernel size $k\times k$ and output size $O\times O$, total multiply-accumulate (MAC) count is $\text{\# MAC} = O^2*N*k^2*M$. 
The total number of $parameters$ for the convolutional layer is given by  $N \times M \times k^2$. We define \textit{OPS reduction} as:
\begin{equation}
    OPS_{reduction} = \frac{(\sum_{i=1}^{L}\text{\# MAC}_i)_{baseline}}{(\sum_{i=1}^{L}\text{\# MAC}_i)_{pruned network}}
\end{equation} 
Our results are summarized in Table \ref{tab1}. The $net 0$ in each case serves as the baseline. We also specify the \textit{`parameters reduction'} computed with respect to the baseline. The best performance of our pruning algorithm was obtained for ResNet18 on CIFAR-10, where we achieved a $\sim200\times$ reduction in parameters and $\sim 67\times$ OPS reduction with $6.2\%$ reduction in accuracy compared to baseline. For TinyImageNet, we achieve $\sim10\times$ and $\sim5\times$ reduction in parameters and OPS for $<1\%$ accuracy loss from baseline. We observe a natural tradeoff between accuracy and OPS reduction as shown in Fig, \ref{accVsOPS} when the pruning criterion ($\rho$) changes across different networks- \textit{net0, net1,..., net3}. We get a very similar curve for parameter reduction as well. For all scenarios in Table \ref{tab1}, the final pruned model chosen is $net 1$ (except CIFAR-10 ResNet18 case for which we choose $net2$). Pruning beyond that (say $net2 ,net 3$ for CIFAR-10) yields a concave $AE$ profile, that is positive slope, (see Fig. \ref{fig_1}) which activates the stopping criterion $\delta$. The tradeoff graph of Fig. \ref{accVsOPS} empirically 
verifies the redundancy intuition that: \textit{A decreasing AE implies we still have some redundancies in the network that can facilitate
pruning without losing accuracy too much. If AE starts increasing, this means that the network has no more
redundancies and pruning further will contribute to drastic accuracy loss.} Further, the tradeoff analysis can be used to heuristically determine a suitable $\rho, \delta$ criteria based on the user's accuracy-energy requirement.  

Among the evaluated datasets, our algorithm produced more accurate ResNet-type models (with lower accuracy loss compared to baseline) than it did for VGG-type models. The compression in terms of OPS reduction achieved with VGG is slightly higher than ResNet (at equivalent pruning levels). For VGG-19 $net 0$ in Fig. 1, the total AE value started around $0.5$ and decreased to approximately $0.3$ over the course of training. For the corresponding ResNet18 network $net 0$ in Fig. \ref{fig_1}, the AE started 10\% higher. We attribute this increased density to the skip connections in the residual network architecture. Throughout the training process, the densities of the ResNet layers maintained a 10\% higher AE over their VGG counterparts. Since the pruned network's size is determined from AE, higher value of AE implies lower network compression which justifies our ResNet vs. VGG results.






\subsection{Training complexity}

Our method trains progressively smaller networks (networks pruned at each saturation point $\rho$) which reduces the overall training complexity, an advantage of pruning in training. We define `training complexity' as:
\begin{equation}\label{training_complexity}
    \sum_{net_{i}} (\text{OPS reduction}_{net_i})^{-1} \times (\text{\# training epochs}_{net_i})
\end{equation}
where $net_i$ is the set of successively pruned networks (i.e. $\{net 0, net 1, ...\}$) for a given starting configuration. In Table \ref{tab1}, we specify the training epochs for each network ($\text{\# training epochs}_{net_i}$) until the saturation or pruning criteria $\rho$ is satisfied. As an example, training complexity for CIFAR-100 is calculated as $25*(\text{OPS reduction})_{net 0}^{-1} + 210*(\text{OPS reduction})_{net 1}^{-1}$.


\begin{table*}[!t]
    \centering
        \caption{Training complexity for our pruning method. Values $<1\times$ denote reduction in training complexity.}
    \begin{tabular}{|c|c|c|c|c|c|}
        \hline
        
         \textbf{Network} & \multicolumn{3}{|c|}{\textbf{ResNet18}} & \multicolumn{2}{|c|}{\textbf{VGG-19}}\\
        \cline{2-6}
        & CIFAR-10 & CIFAR-100 & Tiny ImNet & CIFAR-10 & CIFAR-100\\
        \hline
        net 0 & 210.0 (1$\times$) & 210.0 (1$\times$) & 60.0 (1$\times$) & 210.0 (1$\times$) & 210.0 (1$\times$) \\
        net 1 & 135.0 (0.64$\times$) & 66.2 (0.32$\times$) & 37.7 (0.62$\times$) & 120.2 (0.57$\times$) & 64.6 (0.31$\times$)\\
        net 2 & 120.8 (0.58$\times$) & - & - & - & - \\
        \hline
    \end{tabular}
    \label{tab3}
    \vspace{-3mm}
\end{table*}
With the training complexity metric, we are essentially measuring the amount of time and training energy required to achieve a given model accuracy, compression (`parameters reduction'), and efficiency (`OPS reduction'). Table \ref{tab3} shows the training complexity of selected networks from our pruning procedure. Note that the networks referred to by Table \ref{tab3} across different datasets are same as that of Table \ref{tab1}. For ResNet18 on CIFAR-10, we see that $net 1$ is objectively better than $net 0$, since it achieved the same accuracy as $net 0$ but with fewer parameters, fewer OPS, and a lower training complexity. Note that, in our evaluation in Eqn. 4, the final pruned network as well as the baseline for CIFAR-10,100 (TinyImageNet) were trained  for 210 (60) epochs, respectively.

\subsection{Visualization}

\begin{figure}
    \centering
    \includegraphics[width=\linewidth]{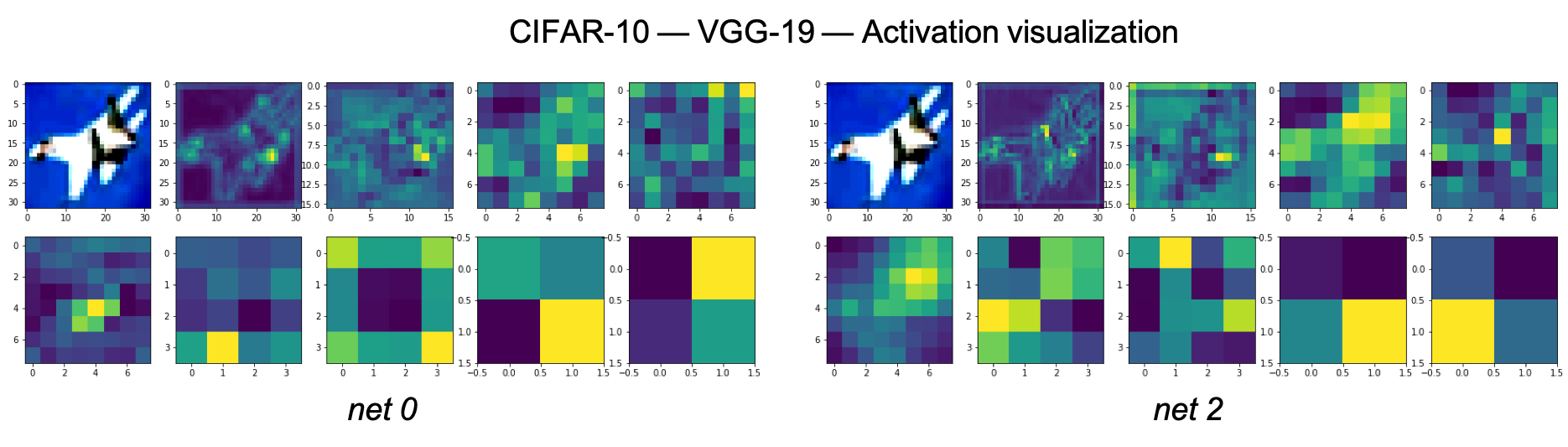}
    \vspace{-6mm}
    \caption{Colormap visualization of the output activations of selected layers in each network. From left to right, top to bottom, the layers represented are: input, 1, 3, 5, 7, 9, 11, 13, 15, 17. For each network and layer, activations from all filters were averaged to produce the colormap shown here.}
    \label{fig_3}
\end{figure}

In addition to plotting the total AE per epoch for each successively pruned network, it proved helpful to visualize the increasing activation density using a colormap. Increasing AE in pruned networks indicates more non-redundancy. The result of this visualization is shown in Fig. \ref{fig_3} for VGG-19 on CIFAR-10. 
Although certain layers break the pattern, we see an overall trend of higher AE (more color implying more neuronal activation) in the layers of $net 2$ than in the layers of $net 0$, the baseline network.

\subsection{Comparison with previous work}\label{compareprevwork}

One of the notable differences of our proposed method against most previously proposed pruning techniques is that it is independent of having pre-trained network initialization. We optimize the network architecture real-time during training which in turn yields additional training complexity reduction. Table \ref{tab2} compares our results with that of two recent works \cite{garg2018low}, \cite{liu2017learning} that aimed to reduce the total number of time intensive pruning-retraining iterations. However, both works still relied on fully or partially trained networks as a starting point.  


\begin{table}[]
    \centering
        \caption{Comparison with previous work for VGG-19 CIFAR-100. The parameters reduction and OPS reduction are with respect to the unpruned VGG-19 baseline.}
    \begin{tabular}{|c|c|c|c|c|}
        \hline
        \textbf{Authors} & \textbf{Training} & \textbf{Accuracy} & \textbf{Parameters} & \textbf{OPS} \\
        & \textbf{complexity} & & \textbf{reduction} & \textbf{reduction} \\
        \hline
        Garg et al. \cite{garg2018low} & 206.6 & 71 \% & 9.1$\times$ & 3.9$\times$\\
        Liu et al. \cite{liu2017learning} & 260.0 & 73 \% & 8.7$\times$ & 1.6$\times$\\
        Ours & 64.6 & 73 \% & 3.9$\times$ & 5.3$\times$\\
        \hline
    \end{tabular}
    \label{tab2}
    \vspace{-3mm}
\end{table}

For example, Liu et al. \cite{liu2017learning} train VGG-19 for 160 epochs, apply their pruning method, then train the pruned network (with an OPS reduction of $1.6\times$ compared to baseline) for another 160 epochs. The initial training adds an unavoidable `160 epochs' of training complexity. Similarly, Garg et al. use PCA on a fully pre-trained network to find the optimal architecture in one single shot. Our method achieves a better training complexity score because we prune \textit{in training}, allowing us to only train the original ($1\times$ OPS reduction) network for 25 epochs before pruning it. When we finally train all the way to a decent accuracy, the OPS in our pruned model have been sufficiently reduced that the final training period of 210 epochs does not incur a heavy training complexity penalty. 

A noteworthy observation here is that our approach yields the highest benefits in terms of OPS reduction while yielding lower parameter reduction savings than Liu et al. and Garg et al. This implies that AE driven pruning methodology prioritizes on pruning the OPS intensive layers more aggressively than the parameter intensive layers.
From Table \ref{tab1} CIFAR-100 VGG-19 results, we observe that on an average $38\%$ of channels are pruned in the first 8 layers $layer1-8$ and $25\%$ channels in the latter 8 layers $layer9-16$. This justifies our above observation and establishes the effectiveness of the AE driven pruning for structured layer-wise network compression focused on overall OPS reduction. 
\begin{table*}[]
    \centering
        \caption{Comparison with Lotter Ticket Hypothesis (LTH) \cite{frankle2018lottery} for CIFAR-10. The parameters reduction and OPS reduction are with respect to the unpruned baseline.}
    \begin{tabular}{|c|c|c|c|c|c|}
        \hline
        \textbf{Model} & \textbf{Authors} & \textbf{Training} & \textbf{Accuracy} & \textbf{Parameters} & \textbf{OPS} \\
        & & \textbf{memory complexity} & & \textbf{reduction} & \textbf{reduction} \\
        \hline
        ResNet18 & LTH \cite{frankle2018lottery} & 206.45 & 93 \% & 5.6$\times$ & N/A\\
        &Ours & 120.8 & 95 \% & 41.2$\times$ & 23.2$\times$\\
        \hline
        VGG-19 & LTH \cite{frankle2018lottery} & 105.1 & 93 \% & 35.7$\times$ & N/A\\
        &Ours & 129.4 & 93 \% & 10.3$\times$ & 27.4$\times$\\
        \hline
    \end{tabular}
    \label{tab4}
    \vspace{-5mm}
\end{table*}

By far, the paper most similar to our work is the notable Lotter Ticket Hypothesis (LTH) \cite{frankle2018lottery}. In \cite{frankle2018lottery}, the authors present the idea of subnetworks that train to comparable degrees of accuracy as their larger parent networks, using iterative magnitude pruning, to find these subnetworks. In essence, our pruning method provides a way to uncover the sub-architectures that train optimally. A key difference between our work and theirs, however, is that we put an emphasis on OPS reduction while they prioritize parameter reduction. Since they do not report OPS counts, it is difficult to compare our work to theirs on the training complexity metric. However, we can show a direct comparison by considering something similar to training complexity, but with parameters instead of OPS factoring into the calculation. We define this quantity as \textit{training memory complexity}: $\sum_{net_{i}} (\text{Parameters reduction}_{net_i})^{-1} \times (\text{\# training epochs}_{net_i})$. Instead of the \textit{energy-complexity} of training as given in Eqn. \ref{training_complexity}, this metric evaluates the \textit{memory-complexity} of training. Table \ref{tab4} shows the comparison.

Our evaluation of LTH takes into account their training methodology as follows: For ResNet18 ({\color{red}VGG-19}) - 20K ({\color{red}10K}) warm up iterations of the unpruned network, then 5K ({\color{red}28K}) iterations with 64.4\% ({\color{red}41\%}) of weights remaining, 5K ({\color{red}28K}) iterations with 41.7\% ({\color{red}16.8\%}), 5K ({\color{red}28K}) iterations with 27.1\% ({\color{red}6.9\%}), and a final 5K ({\color{red}28K}) iterations with 17.8\% ({\color{red}2.8\%}) of weights remaining (their final pruning reduction). With a batch size of 128, the number of training epochs is given by $\frac{\# iterations \times 128}{\#Training Data Size}$. As seen in Table \ref{tab4}, our method outperforms LTH on ResNet18 across all metrics. In contrast, for VGG-19, at iso-accuracy, LTH yields lower parameters and training memory complexity than our method. The authors of LTH acknowledge the limits of their algorithm with regards to scaling on large datasets: they say,``iterative pruning is computationally intensive, requiring training a network 15 or more times consecutively for multiple trials." In contrast, our algorithm finds optimal sub-architectures in a single trial, allowing for use on large datasets such as CIFAR-100, TinyImageNet. As future work, we intend to explore some combination of our work and LTH to yield optimal performance across all network architectures and datasets.

\section{Conclusion}
We propose a novel pruning in training method that yields significant compression benefits on state-of-the-art deep learning architectures. To conduct structured layer-wise pruning, we propose an `Activation Density' metric, a simple yet powerful heuristic that provides a structured and visually interpretable way of optimizing the network architecture. Furthermore, the progressive downsizing of a network during the training process yields training benefits. We get considerable benefits in training complexity and compute-OPS-reduction over the baseline unpruned model, as well as over previously proposed pruning methods. 

Finally, we would like to consider other ramifications of our technique. In essence, our method penalizes networks for having zeros in their activations, forcing the pruned network to have a denser set of activations in comparison to the baseline. A possible negative consequence of enforcing higher density without zeros for ReLU-based networks is that they will not be able to learn non-linearities due to the linear profile of the ReLU function in the regime of positive-only inputs. In the limiting case where a network becomes 100\% dense on a given dataset---where there is never an activation that is zero---the network will be over-fitted and have a greatly reduced ability to generalize on test data. However, the existence of an activation energy saturation point $\rho$ implies that it will be very difficult to create such a 100\% dense network.

\section*{Acknowledgment}
This work was supported in part by National Science Foundation (Grant\#1947826), and the Amazon Research Award.




%
\bibliographystyle{IEEEtran}
\bibliography{biblo}

\end{document}